# A Deeper Look at Experience Replay

**Shangtong Zhang, Richard S. Sutton**
Dept. of Computing Science
University of Alberta
{shangtong.zhang, rsutton}@ualberta.ca

## Abstract

Recently experience replay is widely used in various deep reinforcement learning (RL) algorithms, in this paper we rethink the utility of experience replay. It introduces a new hyper-parameter, the memory buffer size, which needs carefully tuning. However unfortunately the importance of this new hyper-parameter has been underestimated in the community for a long time. In this paper we did a systematic empirical study of experience replay under various function representations. We showcase that a large replay buffer can significantly hurt the performance. Moreover, we propose a simple $\mathcal{O}(1)$ method to remedy the negative influence of a large replay buffer. We showcase its utility in both simple grid world and challenging domains like Atari games.

## 1 Introduction

Experience replay enjoys a great success recently in the deep RL community and has become a new norm in many deep RL algorithms (Lillicrap et al. 2015; Andrychowicz et al. 2017). Until now it is the only method that can generate uncorrelated data for the online training of deep RL systems except the use of multiple workers (Mnih et al. 2016), which unfortunately changes the problem setting somehow. In this paper, we rethink the utility of experience replay. Some critical flaw of experience replay is hidden by the complexity of the deep RL systems, which explains the confusing phenomena that experience replay itself was proposed in the early age of RL, but it did not draw much attention when tabular methods and linear function approximation dominated the field. Experience replay not only provides uncorrelated data to train a neural network, but also significantly improves the data efficiency (Lin 1992; Wang et al. 2016), which is a desired property for many RL algorithms as they are often pretty hungry for data. Although algorithms in pre-deep-RL era do not need to care about how to stabilize a neural network, they do care data efficiency. If experience replay is a perfect idea, it should already be widely used in early ages. However unfortunately, to our best knowledge no previous work has shown what is wrong with experience replay.

Moreover, with the success of the Deep-Q-Network (DQN, Mnih et al. 2015), the community seems to have a default value for the size of the replay buffer, i.e. $10^6$. For instance, Mnih et al. (2015) set the size of their replay buffer for DQN to $10^6$ for various Atari games (Bellemare et al. 2013), after which Lillicrap et al. (2015) also set their replay buffer for Deep Deterministic Policy Gradient (DDPG) to $10^6$ to address various Mujoco tasks (Todorov et al. 2012). Moreover, Andrychowicz et al. (2017) set their replay buffer to $10^6$ in their Hindsight Experience Replay (HER) for a physical robot arm and Tassa et al. (2018) use a replay buffer with capacity as $10^6$ to solve the tasks in the DeepMind Control Suite. In the aforementioned works, the tasks vary from simulation environments to real world robots and the function approximators vary from shallow fully-connected networks to deep convolutional networks. However they all use a replay buffer with same capacity. It seems to be robust under complex deep RL systems and nobody bothers to tune the replay buffer size. However when we separate the experience replay from complex learning systems, we can easily find the agent is pretty sensitive to the size of the replay buffer. Some facts of replay buffer size are hidden by the complexity of the learning system.

Our first contribution is that we present a systematic evaluation of experience replay under various function representations, i.e. tabular case, linear-function approximation and non-linear function approximation. We showcase that both a small replay buffer and a large replay buffer can heavily hurt the learning process. In other words, the size of the replay buffer, which has been under-estimated by the community for a long time, is an important task-dependent hyper-parameter that needs careful tuning. Some facts of experience replay are hidden by the complex modern deep RL systems.

Another contribution is that we propose a simple method to remedy the negative influence of a large replay buffer, which requires only $\mathcal{O}(1)$ extra computation. To be more specific, whenever we sample a batch of transitions, we add the latest transition to the batch and use the corrected batch to train the agent. We refer to this method as combined experience replay (CER) in the rest of this paper.

It is important to note that experience replay itself is not a complete learning algorithm, it has to be combined with other algorithms to form a complete learning system. In our evaluation, we consider the combination of experience replay with Q-learning (Watkins 1989), following the DQN paradigm.

We perform our evaluation and showcase the utility of CER in both small toy task, e.g. Grid World, and large scale challenging domains, e.g. the Lunar Lander and Atari games.

## 2 Related Work

CER can be treated inaccurately as a special case of prioritized experience replay (PER, Schaul et al. 2015). In PER, Schaul et al. (2015) proposed to give the latest transition a largest priority. However PER is still a stochastic replay method, which means giving the latest transition a largest priority does not guarantee it will be replayed immediately. Moreover, it is important to note that PER and CER are aimed to solve different problems, i.e. CER is designed to remedy the negative influence of a large replay buffer while PER is designed to replay the transitions in the buffer more efficiently. To be more specific, if the replay buffer size is set properly, we do not expect CER can further improve the performance however PER is always expected to improve the performance. Although there is a similar part to CER in PER, i.e. giving a largest priority to the latest transition, PER never shows how that part interacts with the size of the replay buffer and whether that part itself can make a significant contribution to the whole learning system. Furthermore, PER is an $\mathcal{O}(\log N)$ algorithm with fancy data structures, e.g. a sum-tree, which significantly prevents it from being widely used. However CER is an $\mathcal{O}(1)$ plug-in, which needs only little extra computation and engineer effort.

Liu and Zou (2017) did a theoretical study on the influence of the size of the replay buffer. However their analytical study only applies to an ordinary differential equation model, and their experiments did not properly handle the episode end by timeout.

Experience replay can be interpreted as a planning method, because it is comparable to Dyna (Sutton 1991) with a look-up table. However, the key difference is that Dyna only samples *states* and *actions*, while experience replay samples *full transitions*, which may be biased and potentially harmful.

There are also successful trials to eliminate experience replay in deep RL. The most famous one is the Asynchronous Advantage Actor-Critic method (Mnih et al. 2016), where experience replay was replaced by parallelized workers. The workers are distributed among processes, and different workers have different random seeds. As a result, the data collected is still uncorrelated.

## 3 Algorithms

Experience replay was first introduced by Lin (1992). The key idea of experience replay is to train the agent with the transitions sampled from the a buffer of previously experienced transitions. A transition is defined to be a quadruple $(s, a, r, s')$, where $s$ is the state, $a$ is the action, $r$ is the received reward after executing the action $a$ in the state $s$ and $s'$ is the next state. At each time step, the current transition is added to the replay buffer and some transitions are sampled from the replay buffer to train the agent. There are various sampling strategies to sample transitions from the replay buffer, among which uniform sampling is the most popular one. Although there is also prioritized sampling (Schaul et al. 2015), where each transition is associated with a priority, it always suffers from $\mathcal{O}(\log N)$ time complexity. So we therefore constrict our evaluation in uniform sampling.

We compared three algorithms: Q-Learning with online transitions (referred to as Online-Q, Algorithm 1), Q-Learning with experience replay (transitions for training only from the buffer, referred to as Buffer-Q, Algorithm 2) and Q-Learning with CER (referred to as Combined-Q, Algorithm 3). Online-Q is the primitive Q-Learning, where the transition at every time step is used to update the value function immediately. Buffer-Q refers to DQN-like Q-Learning, where the current transition is not used to update the value function immediately. Instead, it is stored into the replay buffer and only the sampled transitions from the replay buffer are used for learning. Combined-Q uses both the current transition and the transitions from the replay buffer to update the value function at every time step.

## 4 Testbeds

We use three tasks to evaluate the aforementioned algorithms: a grid world, the Lunar Lander and the Atari

**Algorithm 1:** Online-Q

Initialize the value function $Q$
**while** *not converged* **do**
    Get the initial state $S$
    **while** *$S$ is not the terminal state* **do**
        Select an action $A$ according to a $\epsilon$-greedy policy derived from $Q$
        Execute the action $A$, get the reward $R$ and the next state $S'$
        Update the value function $Q$ with $(S, A, R, S')$
        $S \leftarrow S'$
    **end**
**end**

**Algorithm 2:** Buffer-Q

Initialize the value function $Q$
Initialize the replay buffer $\mathcal{M}$
**while** *not converged* **do**
    Get the initial state $S$
    **while** *$S$ is not the terminal state* **do**
        Select an action $A$ according to a $\epsilon$-greedy policy derived from $Q$
        Execute the action $A$, get the reward $R$ and the next state $S'$
        Store the transition $(S, A, R, S')$ into the replay buffer $\mathcal{M}$
        Sample a batch of transitions $\mathcal{B}$ from $\mathcal{M}$
        Update the value function $Q$ with $\mathcal{B}$
        $S \leftarrow S'$
    **end**
**end**

**Algorithm 3:** Combined-Q

Initialize the value function $Q$
Initialize the replay buffer $\mathcal{M}$
**while** *not converged* **do**
    Get the initial state $S$
    **while** *$S$ is not the terminal state* **do**
        Select an action $A$ according to a $\epsilon$-greedy policy derived from $Q$
        Execute the action $A$, get the reward $R$ and the next state $S'$
        Store the transition $t = (S, A, R, S')$ into the replay buffer $\mathcal{M}$
        Sample a batch of transitions $\mathcal{B}$ from $\mathcal{M}$
        Update the value function $Q$ with $\mathcal{B}$ and $t$
        $S \leftarrow S'$
    **end**
**end**

game Pong. Figure 1 elaborates the tasks.

Our first task is a grid world, the agent is placed at the same location at the beginning of each episode (*S* in Figure 1(a)), and the location of the *goal* is fixed (*G* in Figure 1(a)). There are four possible actions {*Left*, *Right*, *Up*, *Down*}, and the reward is −1 at every time step, implying the agent should learn to reach the *goal* as soon as possible. Some fixed walls are placed in the grid worlds, and if the agent bumps into the wall, it will remain in the same position.

Our second task is the Lunar Lander task in Box2D (Catto (2011)). The state space is $\mathbb{R}^8$ with each dimension unbounded. This task has four discrete actions. Solving the Lunar Lander task needs careful exploration. Negative rewards are constantly given during the landing, so the algorithm can easily get trapped in a local minima, where it avoids negative rewards by doing nothing after certain steps until timeout.

The last task is the Atari game Pong. It is important to note that our evaluation is aimed to study the idea of experience replay. We are not going to study how the experience replay interacts with a deep convolutional network. To this end, it is better to use an accurate state representation of the game rather than try to learn the representation during an end-to-end training. We therefore use the ram of the game as the state rather than the raw pixels. A state is then a vector in $\{0, \ldots, 255\}^{128}$. We normalize each element of this vector into $[0, 1]$ by dividing 255. The game Pong has six discrete actions.

To conduct experiments efficiently, we introduce timeout in our tasks. In other words, an episode will ends automatically after certain time steps. Timeout is necessary in practice, otherwise an episode can be arbitrarily long. However we have to note that timeout makes the environment non-stationary. To reduce the influence of timeout on our experimental results, we manually selected a large enough timeout for each task, so that an episode rarely ends due to timeout. We set timeout to 5,000, 1,000 and 10,000 for the grid world, the Lunar Lander and the game Pong respectively. Furthermore, we use the partial-episode-bootstrap (PEB) technique introduced by Pardo et al. 2017, where we continue bootstrapping from the next state during the training when the episode ends due to timeout. Pardo et al. 2017 shows PEB significantly reduces the negative influence of the timeout mechanism.

Different mini-batch size has different computation complexity, as a result, throughout our evaluation we do not vary the batch size and use a mini-batch of fixed size 10 for all the tasks. In other words, we sample 10 transitions from the replay buffer at each time step. For CER, we only sample 9 transitions, and the mini-batch consists of the sampled 9 transitions and the latest transition. The behavior policy is a $\epsilon$-greedy policy with $\epsilon = 0.1$. We plot the on-line training progression for each experiment, in other words, we plot the episode return against the number of training episodes during the on-line training.

## 5 Evaluation

### 5.1 Tabular Function Representation

Among the three tasks, only the grid world is compatible with tabular methods.

In the tabular methods, the value function $q$ is represented by a look-up table. The initial values for all state-action pairs are set to 0, which is an optimistic initialization (Sutton (1996)) to encourage exploration. The discount factor is 1.0, and the learning rate is 0.1.

Figures 2 (a - c) show the training progression of different algorithms with different replay buffer size for the grid world task. In Figure 2(a), Online-Q solves the task in about 1,000 episodes. In Figure 2(b), although all the Buffer-Q agents with various replay buffer size tend to find the solution, it is interesting to see that smallest replay buffer works best in terms of both the learning speed and the final performance. When we increase the buffer size from $10^2$ to $10^5$, the learning speed keeps decreasing. When we keep increasing the buffer size to $10^6$, the learning speed catches up but is still slower than buffer size $10^2$. We do not keep increasing the replay buffer size to a larger value than $10^6$ as in all of our experiments the total training steps is less than $10^6$. Things are different in Figure 2(c), all of the Combined-Q agents with different buffer size learn to solve the task at similar speed. When we zoom in, we can find the agents with large replay buffer learn fastest as suggested by the purple line and the yellow line. This is contrary to what we observed with the Buffer-Q agents. From Figure 2(b), we can learn that in the original experience replay a large replay buffer hurts the performance, and through Figure 2(c) it is clear that CER makes the agent less sensitive to the replay buffer size.

Q-learning with a tabular function representation is guaranteed to converge under any data distribution only if each state-action pair is visited infinitely many times (together with some other weak conditions). However the data distribution does influence the convergence speed. In the original experience replay, if a large replay buffer is used, a rare on-line transition is likely to influence the agent later compared with a small replay buffer. We use a simple example to show this. Assume we have a re-play buffer with size $m$, and we sample 1 transition from the replay buffer per time step. We assume the replay buffer is full at current time step and a new transition $t$ comes. We then remove the oldest transition in the replay buffer and add $t$ into the buffer. The probability that $t$ is replayed within $k$ ($k <= m$) time steps is

$$1 - (1 - \frac{1}{m})^k$$

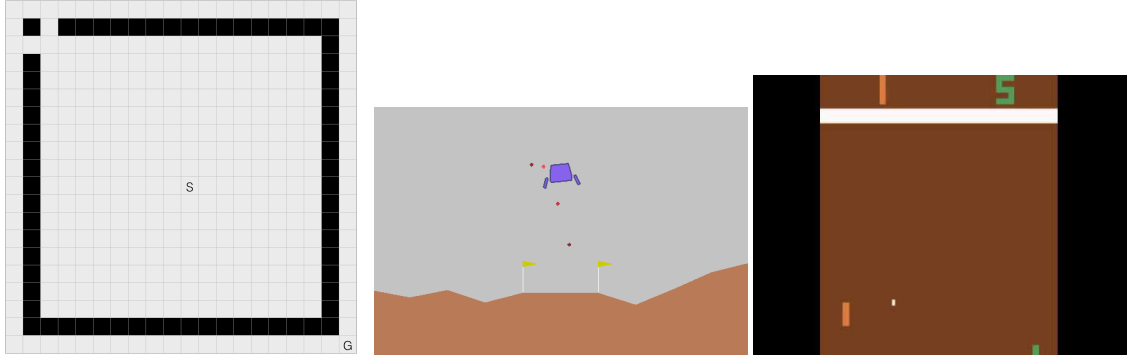

Figure 1: From left to right: the grid world, Lunar Lander, Pong

This function is monotonically decreasing as $m$ increases. So with a larger replay buffer, a rare transition is likely to make influence later. If that transition happens to be important, it will further influence the data collection of the agent in the future. As a result, the overall learning speed is slowed down. This explains the phenomena in Figure 2(b) that when we increase the replay buffer size from $10^2$ to $10^6$ the learning is slowed down. Note with the replay buffer size $10^7$, the replay buffer never gets full thus all transitions are well preserved. It is a special case.

In CER, all the transitions influence the agent immediately. As a result, the agent becomes less sensitive to the selection of the replay buffer size.

## 5.2 Linear Function Approximation

We consider linear function approximation with tile coding (Sutton and Barto (1998)). Among our three tasks, only the Lunar Lander task is compatible with tile coding, so we only consider this task in this part of our evaluation. In our experiments, tile coding is done via the tile coding software [1] with 8 tilings. We set the initial weight parameters to 0 to encourage exploration. The discount factor is to 1.0, and the learning rate is $0.1/8 = 0.125$. The results are summarized in Figure 3. Figure 3(b) shows that a larger replay buffer hurts the learning speed in Buffer-Q. Compared with Figure 3(c), it is clear that adding the on-line transition significantly improves the learning speed, especially for a large replay buffer. The results are similar to what we observed with tabular function representation.

[1] http://incompleteideas.net/sutton/tiles/tiles3.html

## 5.3 Non-linear Function Approximation

We use a single hidden layer network as our non-linear function approximator. We apply the *Relu* nonlinearity over the hidden units, and the output units are linear to produce the state-action value. With a neural network as the function approximator, Buffer-Q is almost the same as DQN. Thus we also employs a target network to gain stable update targets following Mnih et al. 2015. Our preliminary experiments show that random exploration at the beginning stage and a decayed exploration rate ($\epsilon$) do not help the learning process in our tasks.

In the grid world task we use 50 hidden units, and for the other tasks we use 100 hidden units. In the grid world task, we use a one-hot vector to encode the current position of the agent. We use a RMSProp optimizer (Tieleman and Hinton (2012)) for all the tasks, while the initial learning rates vary among tasks. We use 0.01, 0.0005 and 0.0025 for the grid world, the Lunar Lander and the game Pong respectively. These initial learning rates are empirically tuned to achieve best performance.

Figure 4 shows the learning progression of the agents with various replay buffer sizes in the grid world task. We observed that the replay buffer based agents with buffer size 100 and the Online-Q agent fails to learn anything. It is expected as in this case the network tends to over-fit recent transitions thus forgets what it has learned from previous transitions. In Figure 4(a), the Buffer-Q agent with replay buffer size $10^4$ learns fast. This is a medium buffer size rather than the smallest replay buffer size as we observed with tabular and linear function representation. We hypothesize that there is a trade-off between the data quality and data correlation. With a smaller replay buffer, data tends to be more fresh however they are highly temporal correlated, while training a neural network often needs i.i.d. data. With a larger replay buffer, the sampled data tends to be uncorrelated, however they are more outdated. The Buffer-Q agent with extreme large replay buffer (e.g., $10^5$ or $10^6$) fails to find the optimal solution. Comparing Figure 4 (a) and (b), it is clear that CER significantly speeds up the learning, especially for a large replay buffer.

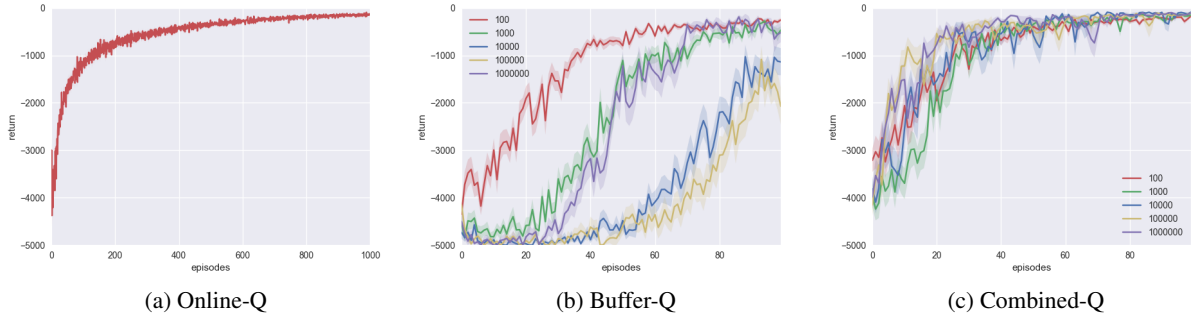

(a) Online-Q  (b) Buffer-Q  (c) Combined-Q

Figure 2: Training progression with tabular function representation in the grid world. Lines with different colors represent replay buffers with different size, and the number inside the image shows the replay buffer size. The results are averaged over 30 independent runs, and standard errors are plotted.

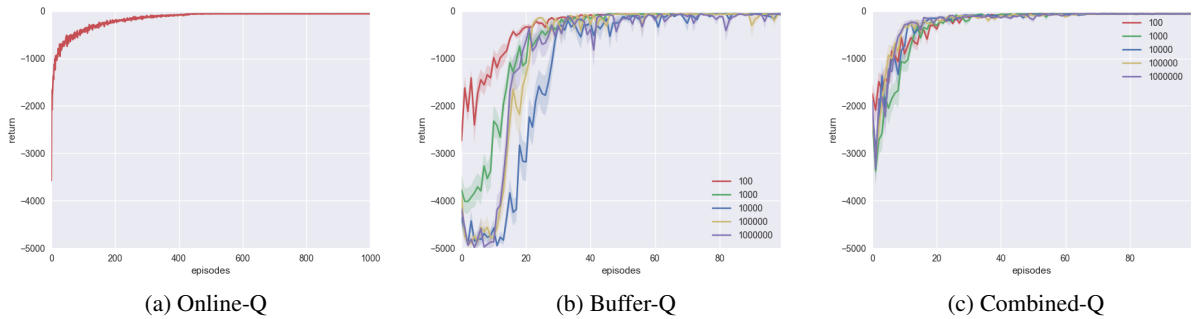

(a) Online-Q  (b) Buffer-Q  (c) Combined-Q

Figure 3: Training progression of linear function approximator on the grid world task. Lines with different colors represent replay buffers with different size, and the number inside the image shows the replay buffer size. The results are averaged over 30 independent runs, and standard errors is plotted.

Figure 5 shows the learning progression of the agents with various replay buffer sizes in the Lunar Lander task. Different from the grid world task, the Online-Q agent and the replay buffer based agents with buffer size 100 do achieve a good performance level. The Online-Q agent achieves almost the best performance among all the agents. This suggests that in this task the neural network function approximator is less likely to over-fit recent transitions. From Figure 5(b), it is clear that the Buffer-Q agent with a medium buffer size ($10^3$) achieves best performance level. With a large replay buffer ($10^5$ or $10^6$), the Buffer-Q agent fails to solve the task. Comparing Figure 5 (b) and (c), we can see that CER does improve the performance for agents with a large replay buffer. One interesting observation is that some replay buffer based agents tend to over-fit the task after certain time steps, thus the performance drops. We found even if we decrease the initial learning rate, this drop still exists.

Figure 6 shows the learning progression of the agents with various replay buffer sizes in the game Pong. We observed similar phenomena as the grid world task. However in this task CER does not provides much improvement.

# 6 Conclusion

Experience replay can improve data efficiency and stabilize the training of a neural network, however it does not come for free. Some important transitions are delayed to make effect by experience replay. This flaw is hidden by the complexity of model deep RL systems. This negative effect is partially controlled by the size of replay buffer, which is shown in this paper to be an important task-dependent hyper-parameter but has been underestimated by the community for a long time. PER is a promising approach addressing this issue, however it often comes with $\mathcal{O}(\log N)$ complexity and non-negligible extra engineer effort. We propose CER, which is similar to a component in PER but only requires $\mathcal{O}(1)$ extra computation, and showcase it can significantly remedy the negative influence of a large replay buffer. However it is important to note that CER is only a workaround, the idea of experience replay itself is heavily flawed. So future effort should focus on developing a new principled algorithm to fully replace experience replay.

**Acknowledgements**

The authors thank Kristopher De Asis and Yi Wan for their thoughtful comments. We also thank Arash Tavakoli, Vitaly Ledvik and Fabio Pardo for pointing out the improper processing of timeout termination in the previous version of the paper.

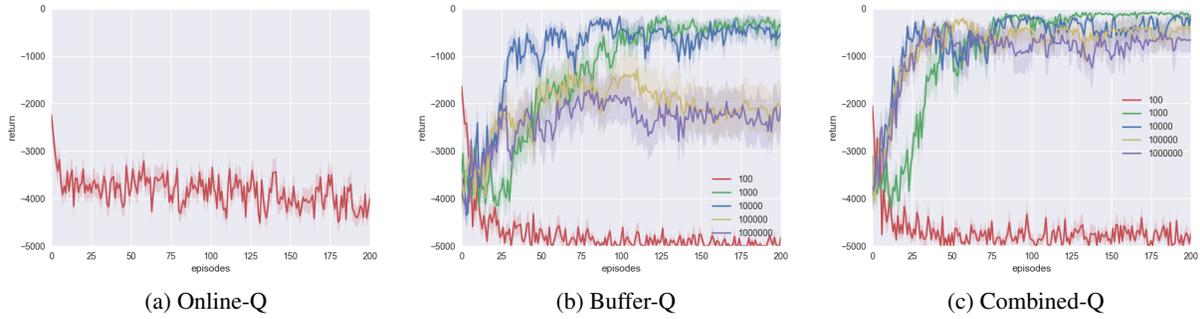

(a) Online-Q  (b) Buffer-Q  (c) Combined-Q

Figure 4: Training progression with non-linear function representation in the grid world. Lines with different colors represent replay buffers with different size, and the number inside the image shows the replay buffer size. The results are averaged over 30 independent runs, and standard errors are plotted.

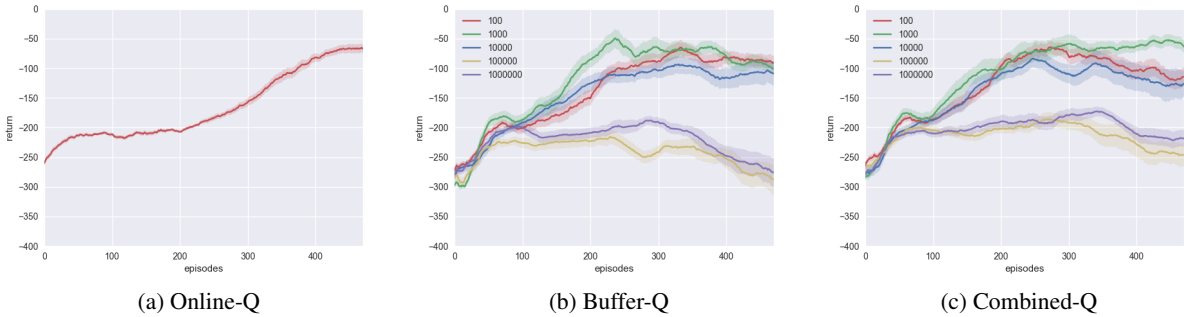

(a) Online-Q  (b) Buffer-Q  (c) Combined-Q

Figure 5: Training progression with non-linear function representation in the Lunar Lander. Lines with different colors represent replay buffers with different size, and the number inside the image shows the replay buffer size. The results are averaged over 30 independent runs, and standard errors are plotted. The curves are smoothed by a sliding window of size 30.

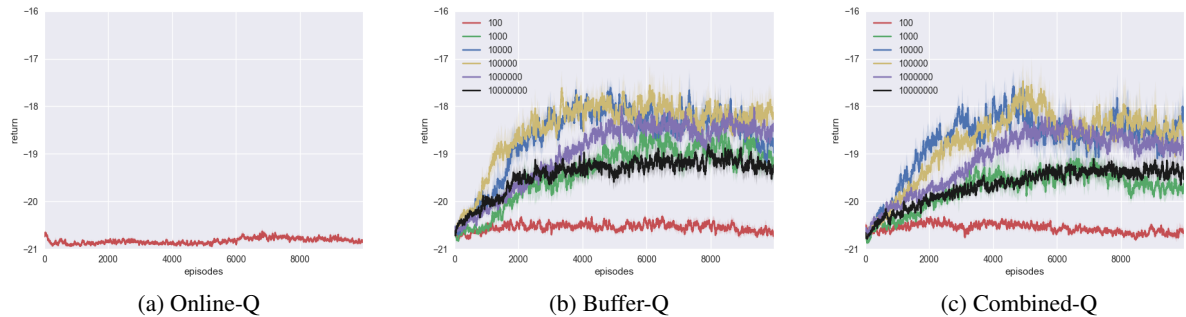

(a) Online-Q  (b) Buffer-Q  (c) Combined-Q

Figure 6: Training progression with non-linear function representation in the game Pong. Lines with different colors represent replay buffers with different size, and the number inside the image shows the replay buffer size. The results are averaged over 10 independent runs, and standard errors are plotted. The curves are smoothed by a sliding window of size 30. It is expected that the agent does not solve the game Pong, as it is to difficult to approximate the state-value function with a single-hidden-layer network.